\documentclass{article}

\usepackage[final]{corl_2019} 

\usepackage[numbers]{natbib}

\usepackage[algoruled,linesnumbered,noend]{algorithm2e}
\usepackage{times}
\usepackage{amsmath}
\usepackage{amssymb}
\usepackage{capt-of}
\usepackage{graphicx}
\usepackage{latexsym}
\usepackage{float}
\usepackage{tabularx}
\usepackage{subcaption}
\usepackage{nicefrac}
\usepackage{bbold}
\usepackage[toc,page]{appendix}
\usepackage{wrapfig}

\usepackage{enumitem}

\title{Disentangled Relational Representations for \\Explaining and Learning from Demonstration}

\author{
    Yordan Hristov\\
    School of Informatics\\
    University of Edinburgh\\
    {\small\texttt{yordan.hristov@ed.ac.uk}}\\
    \And
    Daniel Angelov\\
    School of Informatics\\
    University of Edinburgh\\
   {\small\texttt{d.angelov@ed.ac.uk}}\\
    \And
    Michael Burke\\
    School of Informatics\\
    University of Edinburgh\\
    {\small\texttt{michael.burke@ed.ac.uk}}\\
    \AND
    Alex Lascarides \\
    School of Informatics\\
    University of Edinburgh \\
    {\small\texttt{alex@inf.ed.ac.uk}}\\
    \And
    Subramanian Ramamoorthy\\
    School of Informatics\\
    University of Edinburgh\\
    {\small\texttt{s.ramamoorthy@ed.ac.uk}}
}

\begin{document}
\maketitle

\vspace{-5mm}
\begin{abstract}
Learning from demonstration is an effective method for human users to instruct desired robot behaviour. However, for most non-trivial tasks of practical interest, efficient learning from demonstration depends crucially on inductive bias in the chosen structure for rewards/costs and policies. We address the case where this inductive bias comes from an exchange with a human user. We propose a method in which a learning agent utilizes the information bottleneck layer of a high-parameter variational neural model, with auxiliary loss terms, in order to ground abstract concepts such as spatial relations. The concepts are referred to in natural language instructions and are manifested in the high-dimensional sensory input stream the agent receives from the world.
We evaluate the properties of the latent space of the learned model in a photorealistic synthetic environment and particularly focus on examining its usability for downstream tasks. Additionally, through a series of controlled table-top manipulation experiments, we demonstrate that the learned manifold can be used to ground demonstrations as symbolic plans, which can then be executed on a PR2 robot.
\end{abstract}

\keywords{human-robot interaction, interpretable symbol grounding, learning from demonstration} 


\section{Introduction}

As an increasing number of robots become deployed in field applications, where they must interact in customized ways with human co-workers, there is a need for these robots to represent and reason about their tasks in ways that accord with corresponding human concepts. Ideally, the human's and robot's conceptualizations of the working environment must be able to align so that the robot can adapt to the specific needs of the user. For example, in a table-top manipulation scenario, in order for the agent to correctly respond to instructions regarding stacking or clustering a set of objects, it should be able to comprehend concepts like an object being \textit{close to} or \textit{on} another one---Figure 1.


This motivates the need for a robot to be able to acquire and tune a domain model via interactions with the human user.
Moreover, people who are not robotics experts find it easier to provide the necessary inductive bias in the form of demonstrations of the task rather than explicit specifications of the same task. It is well understood that reward specification is not only hard, but prone to exploitation by the agent \cite{amodei2016concrete}. 
We can therefore use a Learning from Demonstration (LfD) \cite{argall2009survey} method, together with providing high-level guidance using language. This guidance is necessarily more abstract than the level of the robot's sensor stream or native action representation. So, we need to induce alternate latent representations from the low-level sensory data, that allow for subsequent tasks to be grounded in this abstracted space.

\begin{figure}[t]
\centering
\includegraphics[width=1\linewidth]{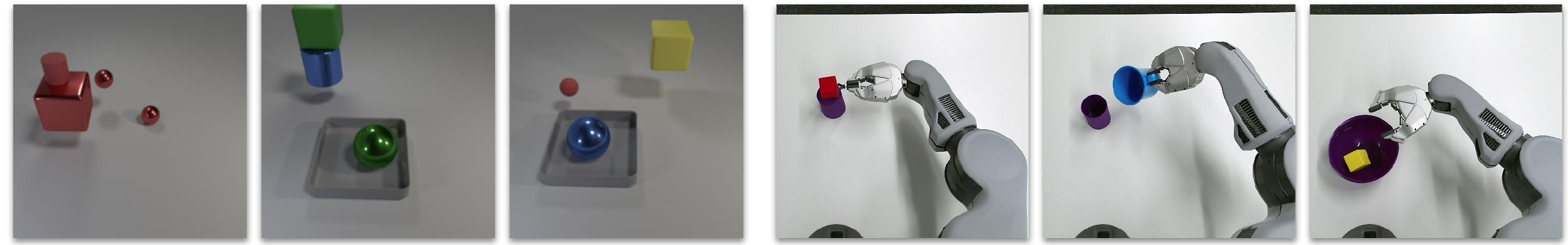}
\caption{Example data used from a \textbf{(a)} photo-realistic blocks world, and \textbf{(b)} table-top object manipulation while teleoperating a 7 DoF arm of a PR2 robot.}
\label{fig:example_data}
\vspace{-3mm}
\end{figure}


Forming a series of hierarchical abstractions about the world that we share with each other---e.g. the notions of color, shape, size, direction, objects' relative position --- is essential for humans to communicate with one another. We would like our robots to also use these human-interpretable concepts as representations that underpin LfD. To achieve this, we work in the setting of interactive task learning \cite{laird2017interactive}, starting with the question of how best to align a learning agent's representations (in this paper, regarding inter-object relationships) with corresponding human labels. 
A specific aspect of this problem is the issue of physical symbol grounding,  \cite{harnad1990symbol, vogt2002physical}, i.e., how should a learning agent make inferences about the {\textit{relationship}} between symbolic labels and their manifestation in the richer sensory feed of the robot.

\begin{figure}[h]
\centering
\includegraphics[width=1\linewidth]{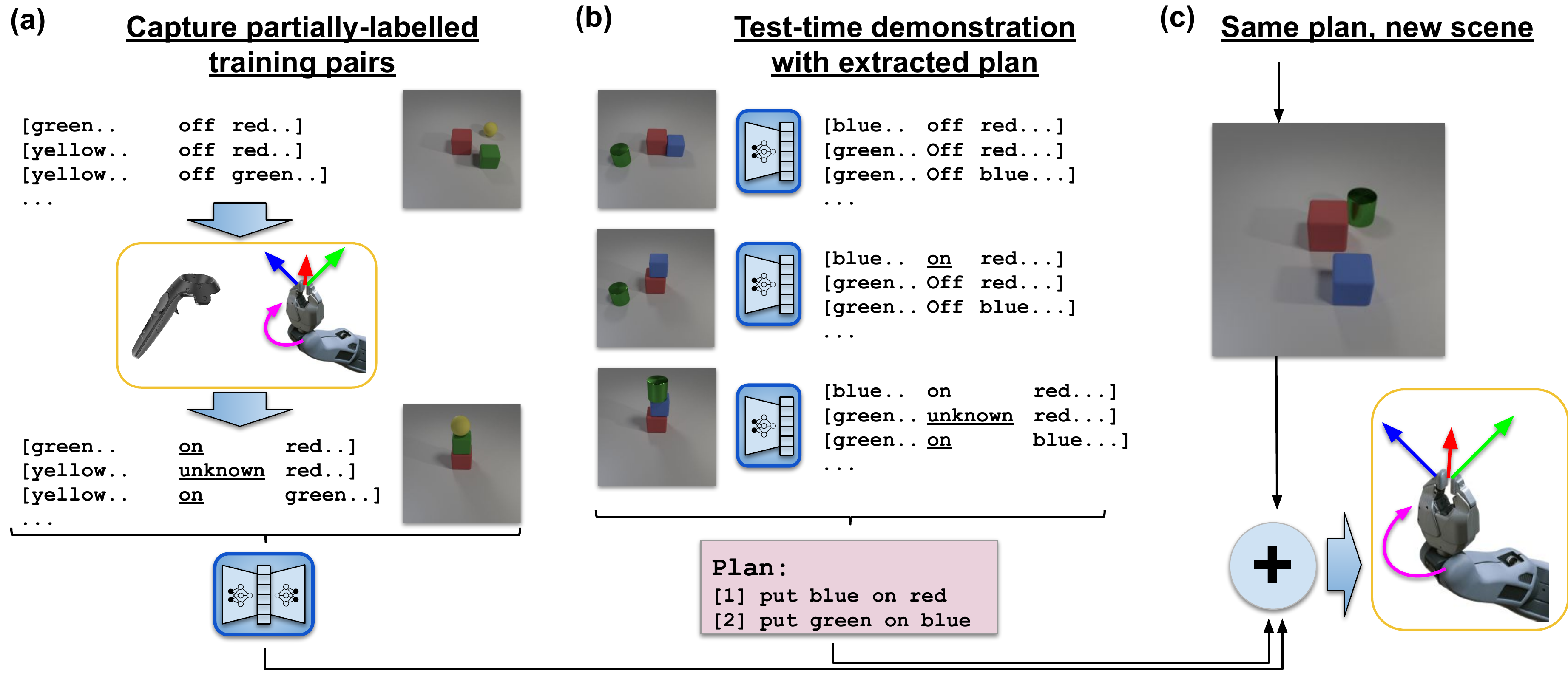}
\caption{Overall setup: \textbf{(a)} during training, the agent receives observations from the environment and weak annotation from the human expert as to how different objects relate to each other, at each time step. \textbf{(b)} At test time, the agent uses the learned representations in order to \textit{explain} how the objects in the environment relate to each other, through time, with the explanation being structured in the form of a plan; \textbf{(c)} each instruction from the plan can then be mapped to end-effector actions.}
\label{fig:explain-n-repeat}
\vspace{-3mm}
\end{figure}

In this paper, we propose a framework which allows human operators to teach a PR2 robot about spatial relations and inter-object arrangements on a table top. Our main contributions are:
\begin{itemize}[topsep=0pt]
    \item A \textbf{disentangled representation} learning method in which inter-object relationships, manifested in a high-dimensional sensory input, can be grounded in a learned low-dimensional latent manifold. We explicitly optimize for the latent manifold to align with human `common sense' notions, e.g. \textit{left} and \textit{right} are mutually exclusive and independent from \textit{front} and \textit{behind} which are also mutually exclusive.
    \item Evaluating the learned representations in an `Explain-n-Repeat' setup---see Figure \ref{fig:explain-n-repeat} (b)---in which \textbf{discrete symbolic specifications}, grounded in the learned manifolds, can be derived from the \textbf{latent projections of user demonstrations}. The demonstrations are third person observations of object manipulation in a table-top environment. We show that we can infer both \textit{what} is moved after what and \textit{how} each object is manipulated from this set of demonstrations. We further demonstrate that end effector poses can be predicted from the steps of such inferred plans, and associated sensory data, see Figure \ref{fig:explain-n-repeat} (c).

\end{itemize}

\section{Related Work}
\label{sec:rel_work}
\vspace{-2mm}

Prior work in psycholinguistics has empirically shown that humans communicate more efficiently and effectively with each other by aligning language and its use at all levels of linguistic processing (e.g., \cite{garrod:anderson:1987,pickering:garrod:2004}). One aspect of the problem is learning how to physically ground symbols in visual input. The INGRESS framework \cite{shridhar2018interactive} uses a multi-step process to learn a representation of objects within the scene, including when  objects are referred to within dialogue with a human.
Learning relationships between objects from raw sensory input can be achieved through the use of high-capacity models like neural networks \cite{santoro2017simple, raposo2017discovering} or with SVMs \cite{rosman2011learning}. However, this can often require large quantities of fully-labelled data and computational resources (e.g., the CLEVR dataset \cite{johnson2017clevr}) and the learned models are often treated as black boxes.

Splitting the factors of variation in an unsupervised way is well studied in the representation learning literature as a form for making the learned models more interpretable. This has been demonstrated using both generative models --InfoGAN \cite{chen2016infogan}, which can be unstable in training and needs specification of the distribution over the latent representation, and variational models of images---$\beta$-VAE \cite{higgins2017beta}, $\beta$-TCVAE \cite{chen2018isolating}, oi-VAE \cite{ainsworth2018interpretable} or of video \cite{denton2017unsupervised}. As these models are trained in an unsupervised way, the resulting embeddings for the factors of variation within the dataset do not necessarily map to the variation that is necessary for the discrimination of the task at hand. In \cite{hristov2018interpretable} the authors employ a $\beta$-VAE representation for grounding of symbols in a semi-supervised way and achieve alignment between the defined semantic concept groups and orthogonal latent vector space representing them. Our work follows this weakly-supervised method of aligning the representations, but differs in that we use the representations to help solve more complex downstream tasks. Moreover, we deal with the segmentation problem when multiple objects are present in the scene.
MONet \cite{burgess2019monet} and IODINE \cite{greff2019multi} present methods for performing iterative multi-object scene decomposition using deep variational inference models. Both approaches choose to solve the scene segmentation and representation learning problems sequentially in an end-to-end fashion, only using unlabelled data. The main focus in both MONet and IODINE is on modelling object-related visual factors of variation. Andreas et al. present Neural Module Netwoks (NMN) \cite{andreas2016neural} which apart from object properties also learn inter-object relations in the context of a Visual Question Answering (VQA) task. However, it is not clear whether what the models learns accords with common-sense human notions. Moreover, it is a fully-supervised approach which might not be a good fit for a realistic LfD setup where explanations for each user demonstration are not expected to be exhaustive.

Lázaro-Gredilla et al. present the Visual Cognitive Computer (VCC) \cite{greff2019multi} and shows how representations that align with human notions and concepts can be learned and then used for a robotic manipulation task. However, the authors assume they have access to a model of the environment and its dynamics, together with a deterministic mapping from sensory inputs to discrete symbols and full plans for each interaction.

On the topic of bridging neural networks and logical plans, Asai et. al \cite{asai2019unsupervised, asai2018classical} present FOSAE - a method for learning how to extract first-order logic predicates and plans from raw sensory observations which can later be composed in a sequential plan. However, the authors claim that the method sacrifices the interpretability of the learned representations for the potential benefit of greater autonomy in the system - which for us is an orthogonal goal, our primary focus being on richer forms of human-robot interaction to help robots acquire customized skills.  
\vspace{-3mm}
\section{Problem Formulation}
\vspace{-1mm}
\label{sec:prob_formulation}



\subsection{Representation Learning Step}

We work with user descriptions which come as natural language sentences of the \verb![target relations referent]! form, where \verb!target! is the object that is manipulated, \verb!referent! is the object that acts as a reference point and \verb!relations! describes the configuration which the \verb!target! should satisfy with respect to \verb!referent!.

Our aim is to efficiently learn how to compress a pair of high-dimensional inputs $I_{tar} \in \mathbb{R}^{D}$, $I_{ref}\in \mathbb{R}^{D}$ to a low-dimensional vector space $\mathbf{C} \subset \mathbb{R}^{L}$, where $L \ll D$, by optimizing a set of functions $q_{\theta} : \mathbb{R}^{D} \xrightarrow{} \mathbb{R}^{Z}$, $q_{\phi} : \mathbb{R}^{Z} \xrightarrow{} \mathbb{R}^{D}$ and $q_{\psi} : \mathbb{R}^{2Z} \xrightarrow{} \mathbf{C} \subset \mathbb{R}^{L}$.

The weak labelling over an observed scene consists of a set of $L$ conceptual groups $\mathcal{G}$ = $\{g_{1}, \ldots ,g_{L} \}$ that aim to describe different notions that are represented in the environment, e.g. alignment along the spatial X/Y/Z axes, containment, support, etc. 
Each group is a set of mutually exclusive discrete labels: $g_{i} = \{y_{1}^{i}, ..., y_{n_i}^{i}\}, n_i = |g_i|$ (e.g. the conceptual group of \textit{alignment along Y} can have the labels \textit{left} and \textit{right}, etc.) Additionally, we have a set of object-centered conceptual groups $\mathcal{O}$ which represent notions like color, shape, size, etc and are extracted from the \verb!target! and \verb!referent! part of the given instructions. Such labels associated with either the target or reference object are designated as $\mathbf{o}_{tar}$ and $\mathbf{o}_{ref}$ respectively.
Let $\mathcal{W} = \{(\mathbf{x}_1, \mathbf{y}_1, \mathbf{o}_{r1}, \mathbf{o}_{t1}), \ldots, (\mathbf{x}_M, \mathbf{y}_M, \mathbf{o}_{rM}, \mathbf{o}_{tM}), (\mathbf{x}_{M+1}, \emptyset, \mathbf{o}_{rM+1}, \mathbf{o}_{tM+1}) \ldots (\mathbf{x}_{N}, \emptyset, \mathbf{o}_{rN}, \mathbf{o}_{tN})\}$ be a set of $N$ observation. $\mathbf{x}_i = (I_{tar}^i, I_{ref}^i)$, $\mathbf{y}_i = \{y^{p} : y^{p} \in g_{p}\}, p \in \{1, \ldots, L\}$; $M$ of the observations are given at least one relational label while the rest are passively gathered as \textit{unknown}. We don't treat the \textit{unknown} value as a label class during training later. 
Each $\mathbf{x}_i$ corresponds to a \verb!(target, referent)! image pair and $\mathbf{y}_i$ corresponds to a \verb!relations! term from the semantically parsed descriptions above. For example, a scene with 3 objects would result in 6 possible bi-object configurations and 6 $(\mathbf{x}, \mathbf{y}, \mathbf{o}_{tar}, \mathbf{o}_{ref})$ pairs respectively. Again note that we expect a proportion of the $\mathbf{y}$ labels to be \textit{unknown}$\equiv$\textit{unlabelled}, due to ambiguity in the scenes, e.g. in Figure \ref{fig:example_data} (second image) the green cylinder is neither left nor right of the blue cylinder. For more details on how linguistic instructions are parsed to labels and how input images are semantically segmented consult Appendix \ref{appendix:architecture}.

We explicitly optimize the vectors in $\mathbf{C}$ to preserve specific semantic concepts expressed over the tuples ($I_{tar}$, $I_{ref}$) and whose meaning is commonly agreed-upon, e.g. relative spatial positions. The latter is achieved by using the vectors in $\mathbf{C}$ to predict the set of labels in each group $g_p \in \mathcal{G}$. Additionally, a subset of the dimensions of each object-centered latent vector $\mathbf{z_i}, i \in \{tar, ref\}$, is forced to predict the values in $\mathbf{o}_{tar}$ and $\mathbf{o}_{ref}$ respectively. 

\subsection{The `Explain-n-Repeat' Step}
\label{sec:explain_n_repeat}
At test time the agent receives a demonstration in the form of a sequence of $T$ raw observations $\mathcal{I} = \{\mathbf{I}_1 \ldots \mathbf{I}_T\}$. For each pair ($\mathbf{o}_{tar}$ and $\mathbf{o}_{ref}$) from the $T$ raw observations we extract a set of semantically-segmented observations $\mathcal{I}_{mask} = \{\mathbf{x}_1 \ldots \mathbf{x}_T\}$ which are projected to a latent embedding trace $\mathcal{T}$. In $\mathcal{T}$ we aim to find a corresponding movement prescription sequence $\mathcal{S}$---which target object moves when---and a sequence of instructions $\mathcal{Y} = \{\mathbf{y}\}$ that is expressed through the symbols that we have learned how to ground in $\mathbf{C}$---how does each target object move. 

To close the loop, when performing the demonstrations on the robot, apart from recording $(\mathbf{x}, \mathbf{y}, \mathbf{o}_{tar}, \mathbf{o}_{ref})$ pairs, we also record the 6 DoF pose $\mathbf{p}$ for the end effector of the arm that is performing the object manipulation. We can thus learn how to regress from an initial image of the scene and a relational specification vector $\mathbf{y}$, describing the end state of the two objects, to a valid pose $\hat{\mathbf{p}}$ which satisfies $\mathbf{y}$. The predicted pose is fed to a MoveIt! motion planner \cite{chitta2012moveit}. We do not address the grasping problem - we assume the robot is already holding the object to be moved.
\section{Methodology}
\label{sec:methodology}

\subsection{Learning Disentangled Relational Embeddings}
\label{sec:embed}
\begin{figure*}[t]
\centering
\includegraphics[width=1\linewidth]{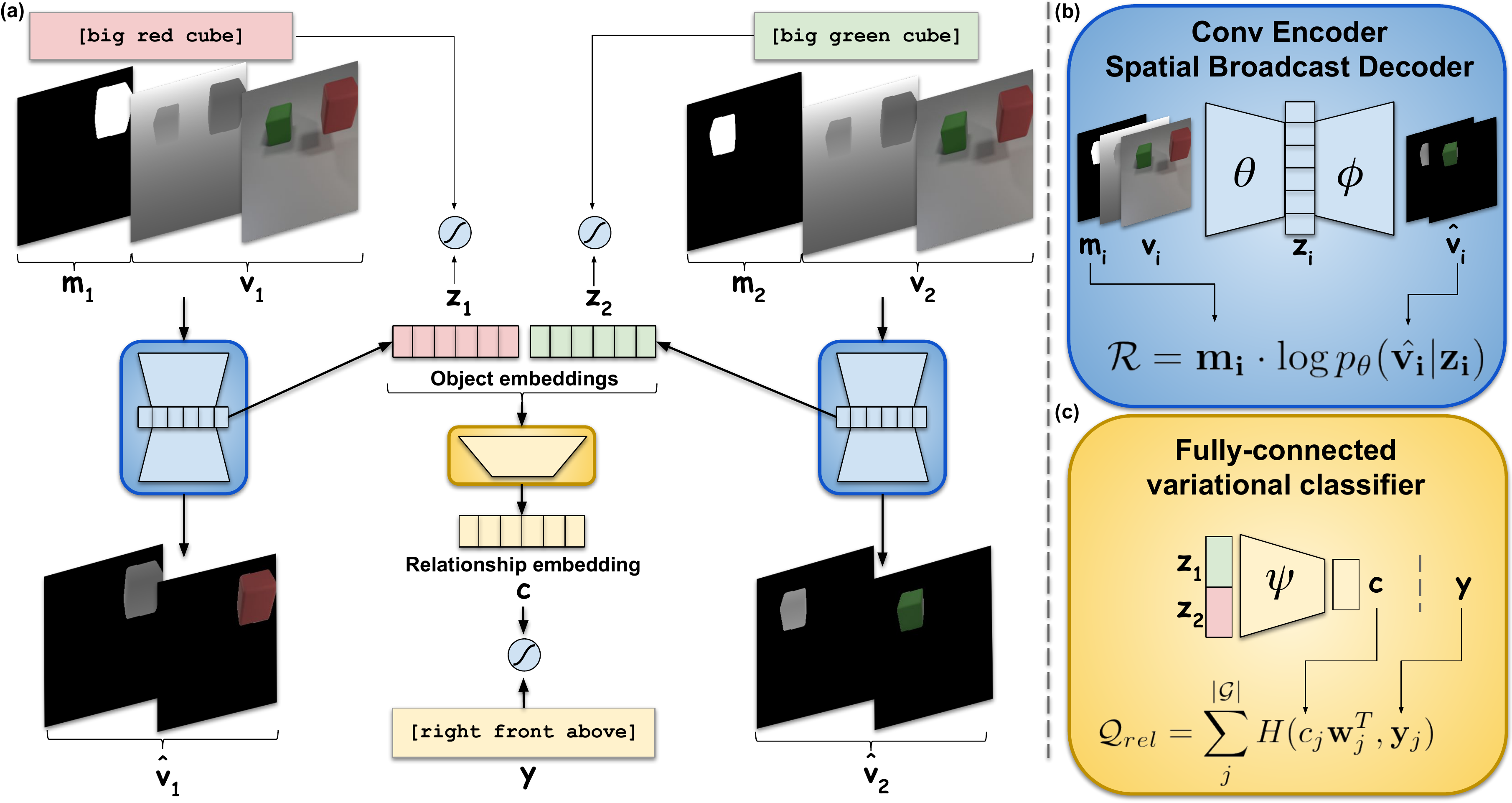}
\caption{\textbf{(a)} Overall architecture - two object-centric embeddings - $\mathbf{z}_1$ and $\mathbf{z}_2$ - are produced for each masked RGBD input - ($\mathbf{m};\mathbf{v}$). From their concatenation a relationship-centric embedding $\mathbf{c}$ is produced. Parts of all embeddings are fed through a set of linear classifiers in order to predict a set of discrete labels - one group of labels per latent axis. Additionally the object-centric embeddings are used to reconstruct the original RGBD inputs $\mathbf{v}$. \textbf{(b)} VAE with a spatial broadcast decoder and masked reconstruction loss, similar to the Component VAE in \cite{burgess2019monet}. \textbf{(c)} Fully connected operator $q_\psi$ for each relational concept group producing a 1D space in which the $\mathbf{y}$ symbols are grounded.}
\label{fig:learning_framework}
\vspace{-4mm}
\end{figure*}

The overall architecture is inspired by the MONet model \cite{burgess2019monet}---augmenting the reconstruction loss term in order to achieve better disentanglement in $\mathbf{Z}$. We do not learn the segmentation process but use already segmented masks. Similar to Hristov et. al \cite{hristov2018interpretable}, we explore the effects of adding auxiliary classification losses to a Siamese Neural Network \cite{koch2015siamese} which uses a $\beta$-VAE \cite{higgins2017beta, burgess2018understanding, kingma2013auto} as a base architecture. It consists of a convolutional encoder network $q_\theta$, parametrized by $\boldsymbol{\theta}$ which takes an input $\mathbf{x}_i$ and produces a vector $\mathbf{z}_i$---red and green object embeddings in Figure \ref{fig:learning_framework} (a). Each $\mathbf{z}_i$ is fed into a spatial broadcast decoder network $p_{\phi}$ \cite{watters2019spatial}, parametrized by $\boldsymbol{\phi}$---Figure \ref{fig:learning_framework} (b).
A set of variational operators $q_\psi$, parametrized by $\boldsymbol{\psi}$, take the concatenation of the $\mathbf{z}$ vectors and produce a single vector $\mathbf{c} \in \mathbf{C}$---yellow relationship embedding in Figure \ref{fig:learning_framework} (b). The resultant vector, $\mathbf{c}$, is fed into a set of linear classifiers $\mathbf{W}$, one per label group $g_i \in \mathcal{G}$, each with a softmax activation function predicting a set of labels. 
Additionally, each $\mathbf{z}_i$ is fed into a set of linear classifiers $\mathbf{W}_o$, one per latent dimension. 

The rationale behind the combination of all of these losses---reconstruction $\mathcal{R}$, variational $\mathcal{KL}$ and multiple classification terms $\mathcal{Q}$---is that they utilize different parts of the dataset in order to achieve the overall goal of learning representations that are factorized and aligned with abstract human notions. The latter is mostly enforced by the Softmax cross-entropy classification terms since they force the latent vectors along each axis to be useful for predicting the labels for a particular concept group. At the same time, the reconstruction loss makes use of all data points, labelled and unlabelled, forcing the same latent vectors to be also useful for recreating the original inputs. As shown in \cite{burgess2019monet}, masking $\mathcal{R}$ forces the encoder network to produce $\mathbf{z}$ which are more factorized.

This, combined with optimizing the Kullback-Leibler divergence between the distribution of values in $\mathbf{C}$ and $\mathbf{Z}$ and a prior isotropic normal distribution, incentivises $\mathbf{C}$ and $\mathbf{Z}$ to be smoother \cite{burgess2018understanding} and for similar data pairs to be projected to the same regions of the manifold.

Additional parameters---$\alpha$ for the reconstruction term, $\beta$ for the Kullback-Leibler divergence term, $\gamma$ for the cross-entropy terms---are used to scale the term in the overall loss---see Equation (\ref{eq:loss}).
\\
\begin{equation}
\begin{aligned}
\min_{\theta, \phi, \psi, \mathbf{W}, \mathbf{W}_o} \mathcal{L}&(\mathbf{x}, \mathbf{y},\mathbf{o} , \theta, \phi, \psi, \mathbf{W}, \mathbf{W}_o) = \beta \mathcal{KL}(\mathbf{C}||\mathbf{Z}) + \alpha \mathcal{R} + \gamma (\mathcal{Q}_{obj} + \mathcal{Q}_{rel}),\\
%
%
&\mathcal{Q} = \mathcal{Q}_{obj} + \mathcal{Q}_{rel} = \sum_{i}^{2}{\sum_{o}^{|\mathcal{O}|} H(z_{io}\mathbf{w}_o^T,\mathbf{o}_{o})} + \sum_{j}^{|\mathcal{G}|} H(q_{\psi}(c_j|\mathbf{z}_1, \mathbf{z}_2)\mathbf{w}_j^T,\mathbf{y}_j)
\end{aligned}
\label{eq:loss}
\end{equation}
In order to evaluate the architecture we perform an ablation study consisting of disabling parts of the full model---e.g. disable the classification part of the network for predicting the object labels and only train the rest. The set of models used in experiments is as follows:\\
    
    \begin{itemize}
    \begin{minipage}{0.49\linewidth}
        \item No $\mathcal{R}$, No $\mathcal{Q}_{obj}$: ($\alpha = 0, \gamma_{obj} = 0$)
        \item No $\mathcal{R}$, With $\mathcal{Q}_{obj}$: ($\alpha = 0, \gamma_{obj} \neq 0$)
    \end{minipage}
    \begin{minipage}{0.49\linewidth}
        \item With $\mathcal{R}$, No $\mathcal{Q}_{obj}$: ($\alpha \neq 0, \gamma_{obj} = 0$)
        \item With $\mathcal{R}$, With $\mathcal{Q}_{obj}$: ($\alpha \neq 0, \gamma_{obj} \neq 0$)
    \end{minipage}
    \end{itemize}

\subsection{Inferring Symbolic Plans from Demonstration}
\label{sec:infer_symbol_plans}
Given the continuous manifold $\mathbf{C}$ in which inter-object relational discrete labels can be grounded, we look into whether that feature space can be used in an LfD context. 
In particular, we investigate whether the learned manifold allows us to segment the latent projections of user demonstrations for moving objects. 

\textbf{Plan segmentation} - Given a sequence of observations $\mathcal{I}$ and reference object labels $\mathbf{o}_{ref}$, a prepossessing step is taken to identify all target objects and to extract masked observations $\mathcal{I}_{mask}$, for each pair of target and reference objects. Each $\mathcal{I}_{mask}$ is projected to a latent projection $\mathcal{T}$ from which a movement prescription sequence $\hat{\mathcal{S}}$ is extracted. The latter designates when an object is manipulated and when not. Using the methods described in Section \ref{sec:embed} we identify the different target (moved) objects---green and blue in Figure \ref{fig:testing_clevr_data} (a). Then for each pair of target and a given reference object---the red cube---we extract the corresponding traces of relational embeddings. Checking whether the particular target object moves with respect to the reference object at each timestep $t$ consists of performing a likelihood ratio test with two candidate normal distributions, parametrized by $\Sigma_{mov}$ and $\Sigma_{stat}$, $\Sigma_{stat} \ll \Sigma_{mov}$. The procedure is more formally described in Algorithm \ref{alg:plan_extraction}, Appendix \ref{appendix:plan_segmentation}. However, in a given set of demonstrations we are not only interested in identifying when one objects stops moving and another starts. We are also interested in how the relationships between them change over time. More specifically, we are interested in being able to identify an invariant symbolic plan $\mathcal{Y}$ that underlies a set of demonstrations, all of which demonstrate the same task.

\textbf{Task essence extraction} - This step is performed in a similar fashion to the plan segmentation step described above. Given N latent projections $\mathcal{T}_1, \ldots, \mathcal{T}_N$ from N demonstrations for a single ($\mathbf{o}_{tar}, \mathbf{o}_{ref}$) pair we use a set of estimated 1D normal distributions for each relational label in each conceptual group: K = $\{\{\mathcal{N}(\mu_{q}^p,\sigma_{q}^p)\}, p \in \{1, \ldots, L\}, q \in \{1, \ldots, |g_p|\}$ to perform label-oriented likelihood ratio tests (as compared to the moving ones in the prev paragraph). As a result each $\mathcal{T}$ is converted to a symbolic trace and the eventual identified plan $\mathcal{Y}_{\mathbf{o}_{tar}}$, for a given $\mathbf{o}_{tar}$, is the most invariant set of symbols from all traces - the task essence. It is worth clarifying that the task essence extraction currently works only for tasks of deterministic nature - there's a single sequence of actions that achieve the goal. For more details refer to the supplementary materials\footnote{https://sites.google.com/view/explain-n-repeat/} or to Appendix \ref{appendix:plan_segmentation}.

\textbf{From symbolic plans to end effector poses} - Predicting end effector poses of the robotic arm is treated as a fully-supervised problem. From an observation of the environment---an image showing a grasped object ($\mathbf{o}_{tar}$) and a static object ($\mathbf{o}_{ref}$) on a table top---we extract the object-centric embedding corresponding to the target object---$\mathbf{z}_{tar}$. Additionally, given a relational vector $\mathbf{y}$, arising from $\mathcal{Y}_{\mathbf{o}_{tar}}$, describing the desired eventual state of the two objects, we sample a relational embedding $\mathbf{c}$, by using the fitted parametric distributions K (see previous paragraph). Given the concatenation of $\mathbf{z}_{tar}$ and $\mathbf{c}$, we use an MLP with two hidden layers in order to regress to a pose vector $\hat{\mathbf{p}} \in \mathbb{R}^6$.
\section{Experiments, Evaluation and Results}
\label{sec:experiments_eval_results}
\vspace{-3mm}

    \begin{wraptable}{l}{0.3\textwidth}
        \centering
        \begin{tabular}{l}
          left, right  \\
          front, behind \\
          above, below  \\
          close, far \\
          on, off  \\
          out of, in \\
          \hline
          off, on  \\
          not facing, facing \\
          out, in  \\
      \end{tabular}
      \caption{User-defined spatial relations}
      \label{table:labels}
      \vspace{-5mm}
    \end{wraptable}
    
    For learning the relational embeddings a set of standard objects is used, as shown in Figure \ref{fig:example_data}. The set of spatial prepositions and their semantic grouping that are given in the user-scene descriptions during the demonstrations are outlined in Table \ref{table:labels}.
    
    \begin{figure*}[t]
    \centering
    \includegraphics[width=1\linewidth]{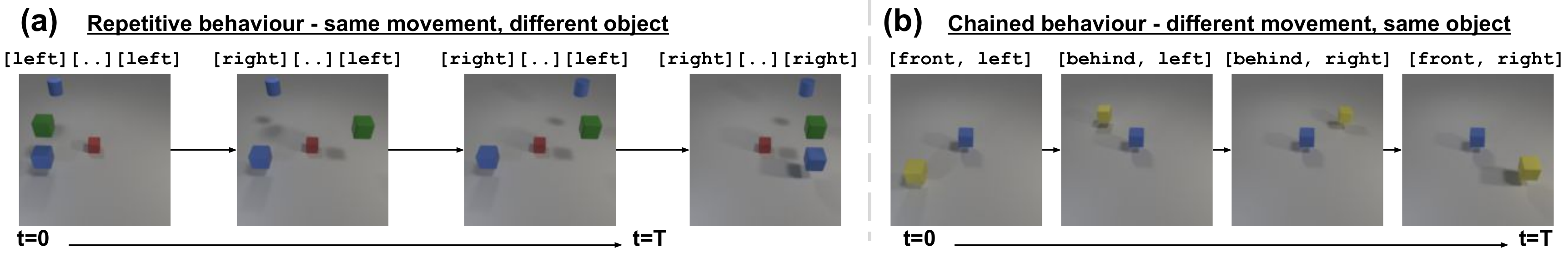}
    \caption{Example testing data for \textbf{(a)} Repetitive motion along a single concept group---e.g. left to right (row 1)---and \textbf{(b)} Chained motion along different concept groups---e.g. perform a C-shape-sequentially from front to behind to right to front (row 1).}
    \label{fig:testing_clevr_data}
    \vspace{-5mm}
    \end{figure*}

    \textbf{Photorealistic BlocksWorld} - This synthetic dataset consists of 1,000 scenes, each containing 4 objects in a random configuration. The objects' attributes are the defaults from the original CLEVR dataset \cite{johnson2017clevr}, together with an additional \textit{gray tray} object. Given the 6 concept groups---Table \ref{table:labels} (top)---this results in 72,000 possible inter-object relationships, 40\% of which are unlabelled. 
    
    It is worth noting that the different concept groups have a different split between labelled and unlabelled data points as an artefact of resolving the inherent ambiguity of some of the prepositions when procedurally generating the different scenes. The decision of whether a relationship is known or not is determined through a set of empirically-defined thresholds whose values are specified before generating the dataset. For example, if an object is above a tray but their vertical distance is less than a threshold the pair is labelled as \textit{unknown} along the in/out concept group.
    The proportion of unlabelled data points across the 6 concept groups is 28\%, 31\%, 41\%, 36\%, 32\%, 90\% respectively.
    
    For evaluating the efficacy of plan segmentation using the learned relation embeddings, two types of moving scenes are generated - 6 \textit{repetitive} behaviours of multiple target objects sequentially moving along a specific concept group (5 demos per type) and 3 \textit{chained} behaviours of the same target object moving along different concept groups (8 demos per type)---see Figure \ref{fig:testing_clevr_data}. Task essence extraction is tested only on the demonstrated chained behaviours. Accuracy is reported for each identified $\hat{\mathcal{S}}$ and edit distance is reported for each symbolic plan---see Equations 2 and \ref{eq:ed}.
    
    \begin{minipage}{.52\linewidth}
      \begin{equation}
       Acc(\mathcal{S}, \hat{\mathcal{S}}) = \frac{1}{T|O_{tar}|} \sum_{j}^{|O_{tar}|} \sum_{i}^{T} \mathbb{1}{(\mathcal{S}_{oi}=\hat{\mathcal{S}}_{oi})}
      \end{equation}
      \label{eq:seg_acc}
    \end{minipage}
    \begin{minipage}{.48\linewidth}
      \begin{equation}
        ed(\mathcal{Y}_o, \hat{\mathcal{Y}_o}) = \frac{1}{|\mathcal{Y}_o|} \sum_{i}^{|\mathcal{Y}_o|} \mathbb{1}{(\mathcal{Y}_{oi}\neq\hat{\mathcal{Y}}_{oi})}
        \label{eq:ed}
      \end{equation}
    \end{minipage}
    
    \textbf{PR2 Robot Experiment} - 3 tasks are demonstrated by teleoperating a PR2 robot with an HTC Vive controller---putting a red cube on a purple cup, making two cups face each other (as an example of a necessary pre-pouring step), placing a yellow cube in a purple bowl---see Figure \ref{fig:example_data} (b). The spatial inter-object prepositions that were learned from each of the 3 tasks are summarized in Table \ref{table:labels} (bottom). The separation of known/unknown depends on the temporal aspect of the demonstrations. For instance, we know that at the beginning and at the end of each demonstration a pair of mutually exclusive relational labels are satisfied, respectively. Everything in the middle is ‘unknown’. Here, what matters is the segmentation of the demonstration into an initial, middle and final stages for which we use a temporal template - first 2s and last 2s correspond to the initial and final stage, the rest is the middle stage. For each task there are 20 demonstrations performed, with variations in the position of the reference object in the scene and initial end effector poses. In total this results in 2,400 labelled and 6,000 unlabelled object pairs.
    
    For evaluating how well we can predict an end effector pose from a given input image and a relational spec vector, we record 10 additional demonstrations for each task. The mean absolute error along each of the 6 axes of the end effector is reported between the inferred set of poses and the ground truth ones, measured in meters for X/Y/Z and radians for Roll/Pitch/Yaw.
\begin{table}[h]
\begin{subtable}{1\linewidth}\centering
{\begin{tabular}{l|rrrrrr}
\hline
Model            & \multicolumn{1}{c}{left-right} & \multicolumn{1}{c}{front-behind} & \multicolumn{1}{c}{below-above} & \multicolumn{1}{c}{far-close} & \multicolumn{1}{l}{off-on} & \multicolumn{1}{l}{out-in} \\ \hline
No $\mathcal{R}$, No $\mathcal{Q}_{obj}$ & 0.50                           & 0.64                             & 0.54                            & 0.56                          & 0.49                       & 0.66                       \\
No $\mathcal{R}$, With $\mathcal{Q}_{obj}$    & 0.53                           & 0.68                             & 0.68                            & 0.63                          & 0.65                       & 0.62                       \\
With $\mathcal{R}$, No $\mathcal{Q}_{obj}$    & 0.70                           & 0.73                             & 0.69                            & 0.68                          & 0.64                       & \textbf{0.78}              \\
With $\mathcal{R}$, With $\mathcal{Q}_{obj}$       & \textbf{0.80}                  & \textbf{0.88}                    & \textbf{0.91}                   & \textbf{0.86}                 & \textbf{0.76}              & 0.56                    
\end{tabular}}
\end{subtable}
	
\begin{subtable}{1\linewidth}\centering
\vspace{3mm}
{\begin{tabular}{l|rrr}
\hline
Model   & \multicolumn{1}{l}{C-shape} & \multicolumn{1}{l}{off-on-off} & \multicolumn{1}{l}{jump over} \\ \hline
All models & 1                           & $\approx$ 0.74                           & 1
\end{tabular}}
\end{subtable}
	\vspace{3mm}
    \caption{Plan segmentation Acc-\textit{what moves when}-for \textbf{(top)} repetitive and \textbf{(bottom)} chained demos.}
\label{table:plan_segmentation_acc}
\vspace{-5mm}
\end{table}

The performed experiments demonstrate that the learned feature space can be reliably used by the agent in order to produce symbolic plans, using the dictionary of symbols it has been taught. Table \ref{table:plan_segmentation_acc} shows that the model which incorporates both $\mathcal{R}$ and $\mathcal{Q}$ performs best at identifying the movement prescription sequences $\hat{\mathcal{S}}$ in the repetitive demonstrations. This supports our hypothesis that by enforcing object label classification and by utilizing the full dataset through the reconstruction loss, we learn smoother and more factorized vectors $\mathbf{z}$ and $\mathbf{c}$. This in turn allows for the task segmentation process to be more robust. Further analysis is provided in Appendix \ref{appendix:violins}.
As far as the chained movement demonstrations are concerned, all models perform in an equal manner, which is expected, since these sequences only involve a single object moving.
\begin{figure*}[h]
    \vspace{-5mm}
    \centering
    \begin{subfigure}[b]{.325\linewidth}
    \centering
    \includegraphics[width=1\linewidth]{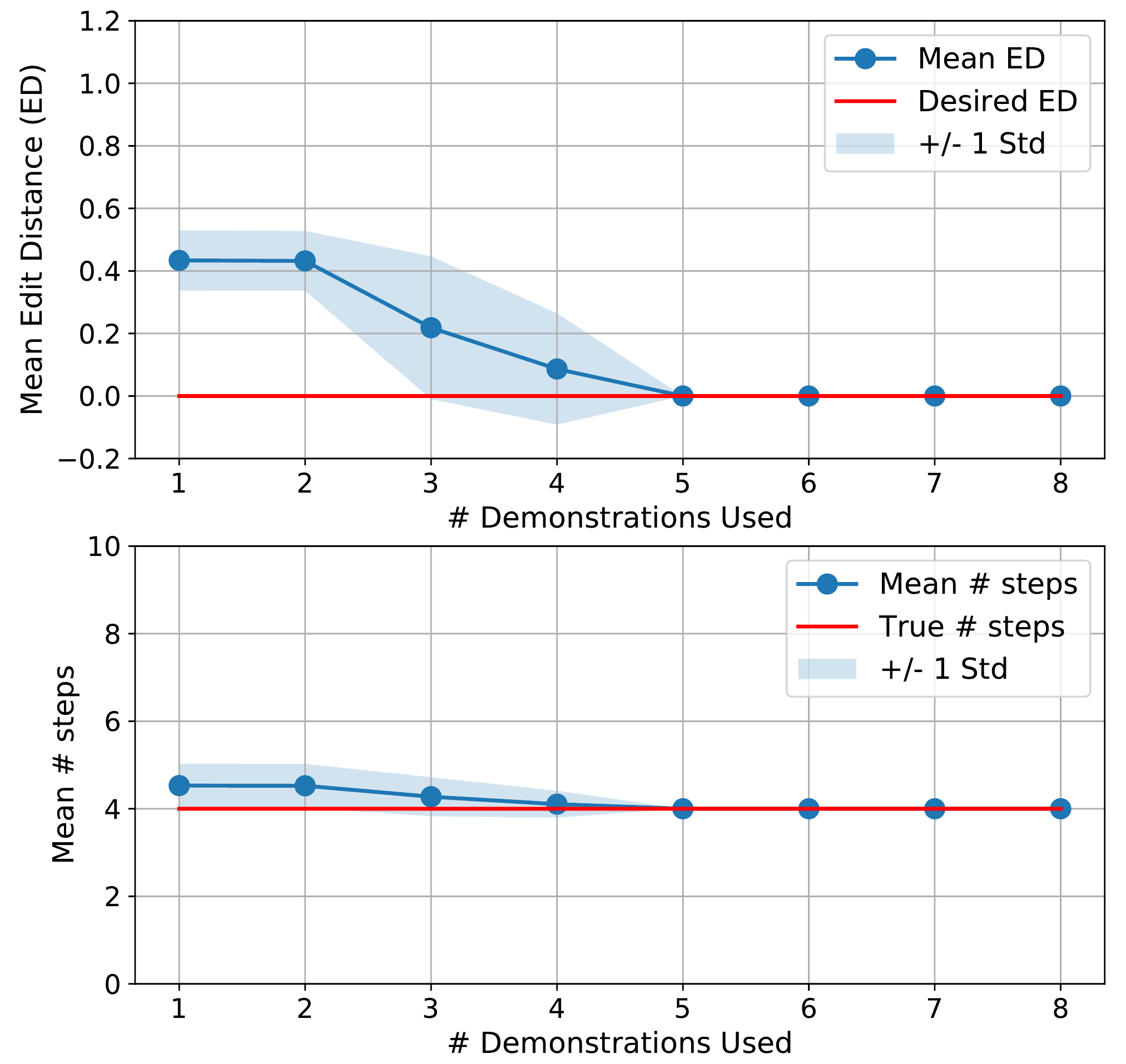}
    \caption{}
    \end{subfigure}\hfill
    \begin{subfigure}[b]{.325\linewidth}
    \centering
    \includegraphics[width=1\linewidth]{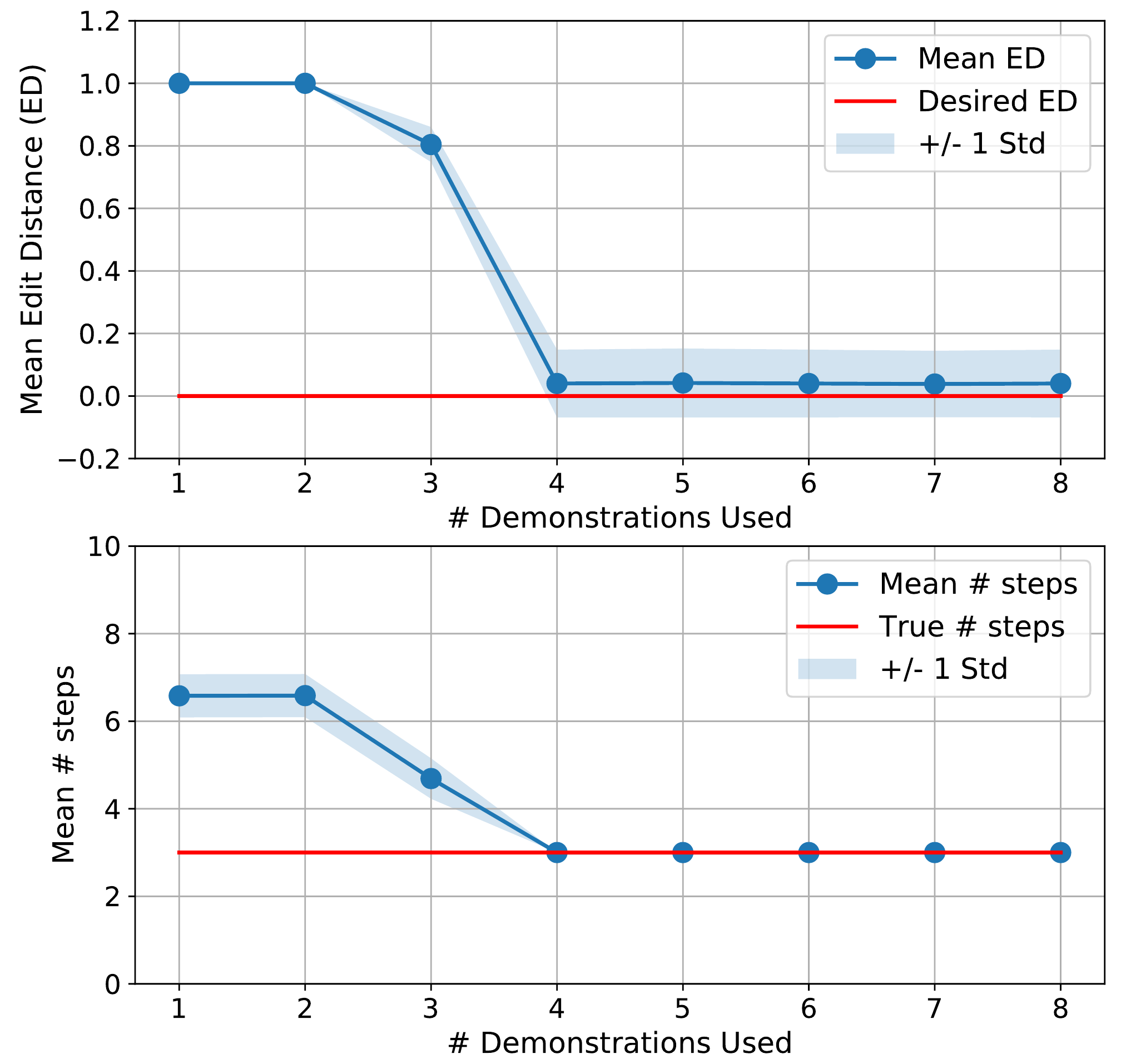}
    \caption{}
    \end{subfigure}\hfill
    \begin{subfigure}[b]{.325\linewidth}
    \centering
    \includegraphics[width=1\linewidth]{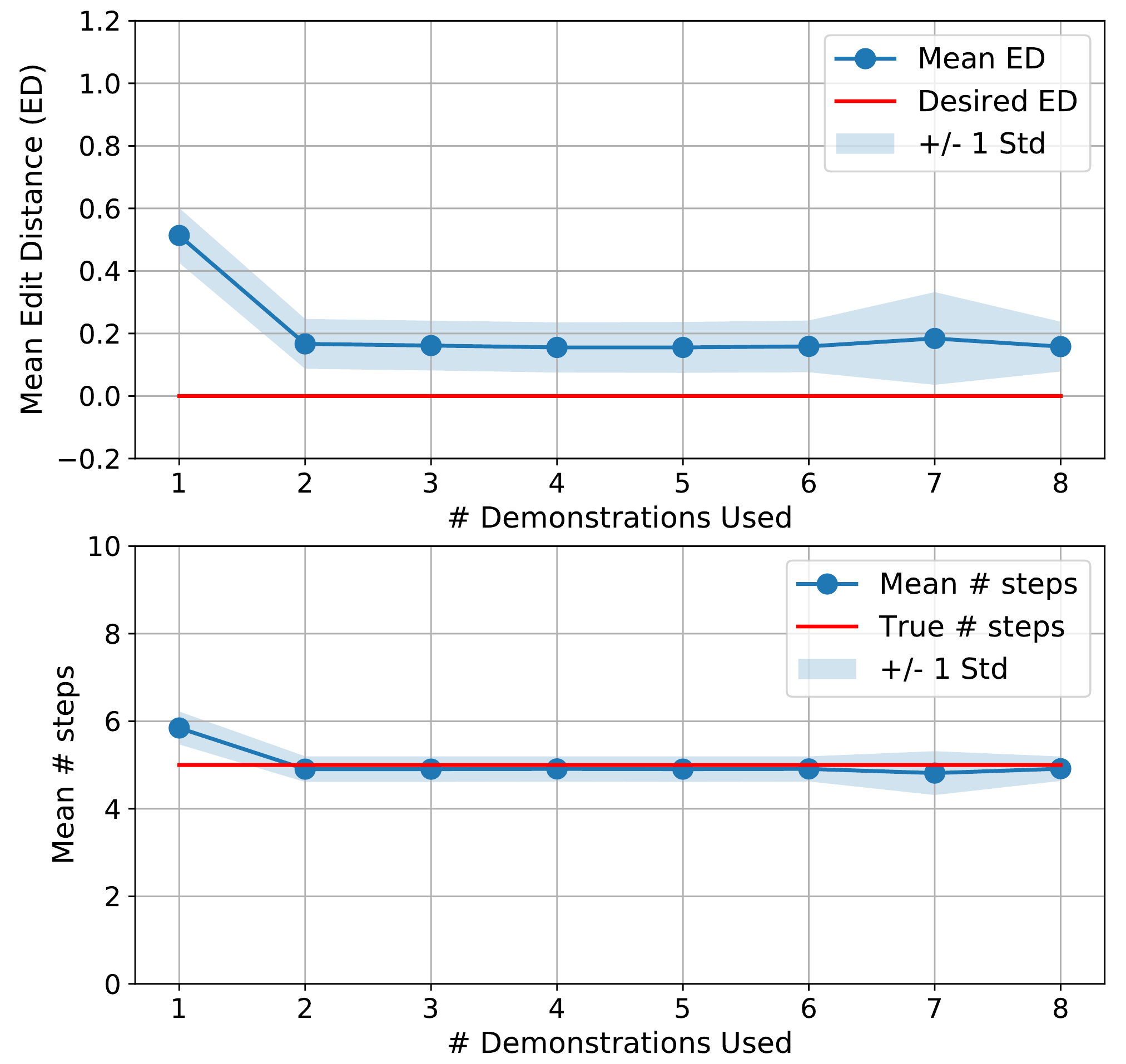}
    \caption{}
    \end{subfigure}
    \vspace{-0.5\baselineskip}
    \caption{(top): edit distance statistics as a function of how many demonstrations the agent has seen. (bottom) plan length statistics for the inferred plans as a function of how many demonstrations the agent has seen for all three chained behaviours---\textbf{(a)} C-shape, \textbf{(b)} off-on-off and \textbf{(c)} jump over; }
    \label{fig:rel_chaining_ed}
    \vspace{-2mm}
\end{figure*}
The best performing model from Table \ref{table:plan_segmentation_acc} is used on the symbolic plan inference task over the demonstrated chained behaviour (where the underlying plan is over a single target object and is a multi-step one). Figure \ref{fig:rel_chaining_ed} reports the average edit distance for the inferred plans $\hat{\mathcal{Y}}$ over all demonstrations for a given task (top row), together with the average plan lengths $|\hat{\mathcal{Y}}|$. Both quantities are plotted as a function of the number of demonstrations used to infer the task essence which in turn is used to infer the step-by-step plan for each demonstration. As expected, the more demonstrations we see per task, the closer the inferred plans $\hat{\mathcal{Y}}$ get to the ground truth ones $\mathcal{Y}$. The reason why some of the plots do not converge to the ground-truth numbers (red line across all plots in the figure) can be attributed to the fact that some demonstrations contain object occlusions, making it hard to reliably infer the true plan without noise. 

\begin{figure*}[h]
    \centering
    \begin{subfigure}[b]{.325\linewidth}
    \centering
    \includegraphics[width=1\linewidth]{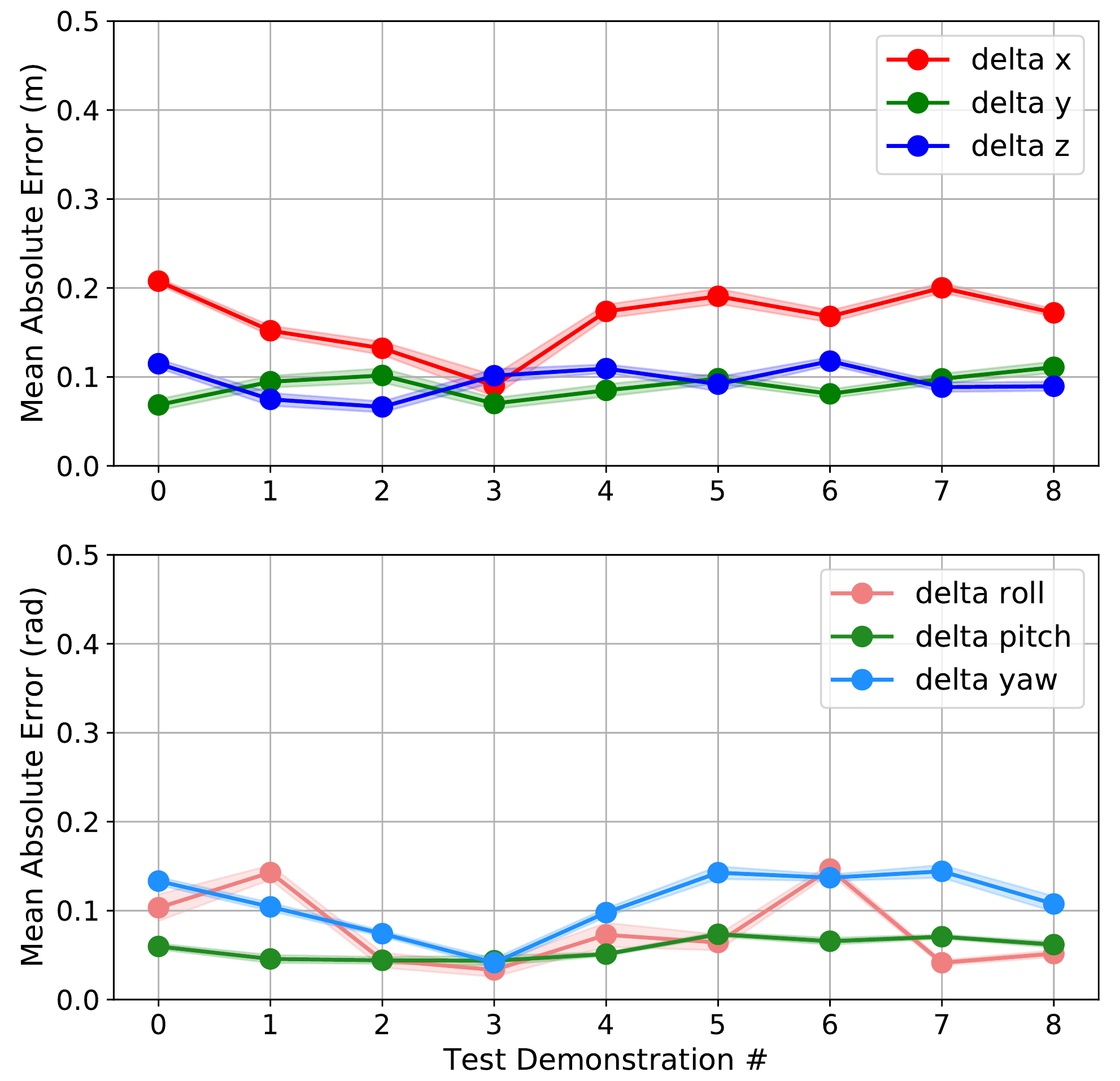}
    \vspace{-1\baselineskip}
    \caption{}
    \end{subfigure}\hfill
    \begin{subfigure}[b]{.325\linewidth}
    \centering
    \includegraphics[width=1\linewidth]{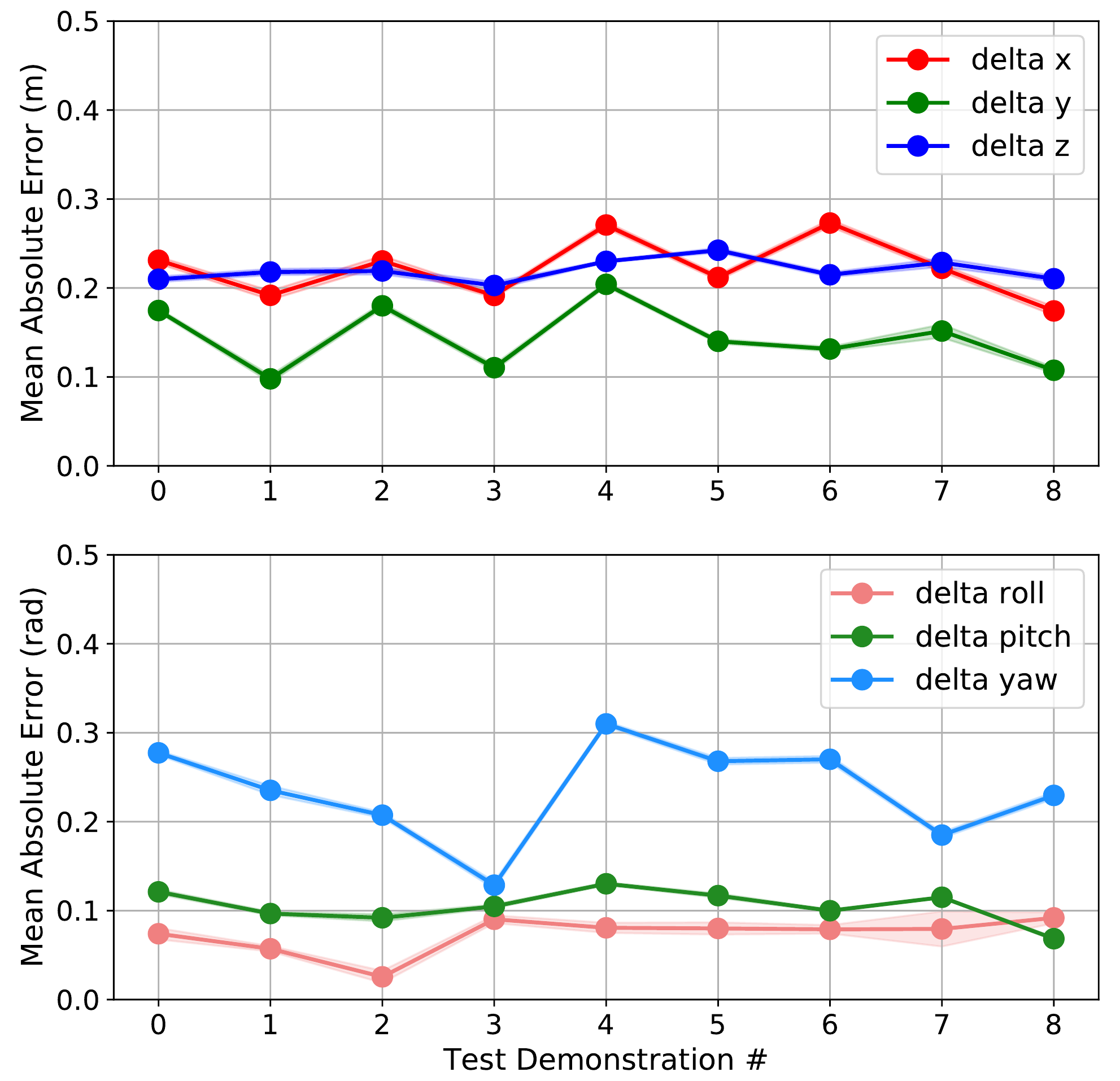}
    \vspace{-1\baselineskip}
    \caption{}
    \end{subfigure}\hfill
    \begin{subfigure}[b]{.325\linewidth}
    \centering
    \includegraphics[width=1\linewidth]{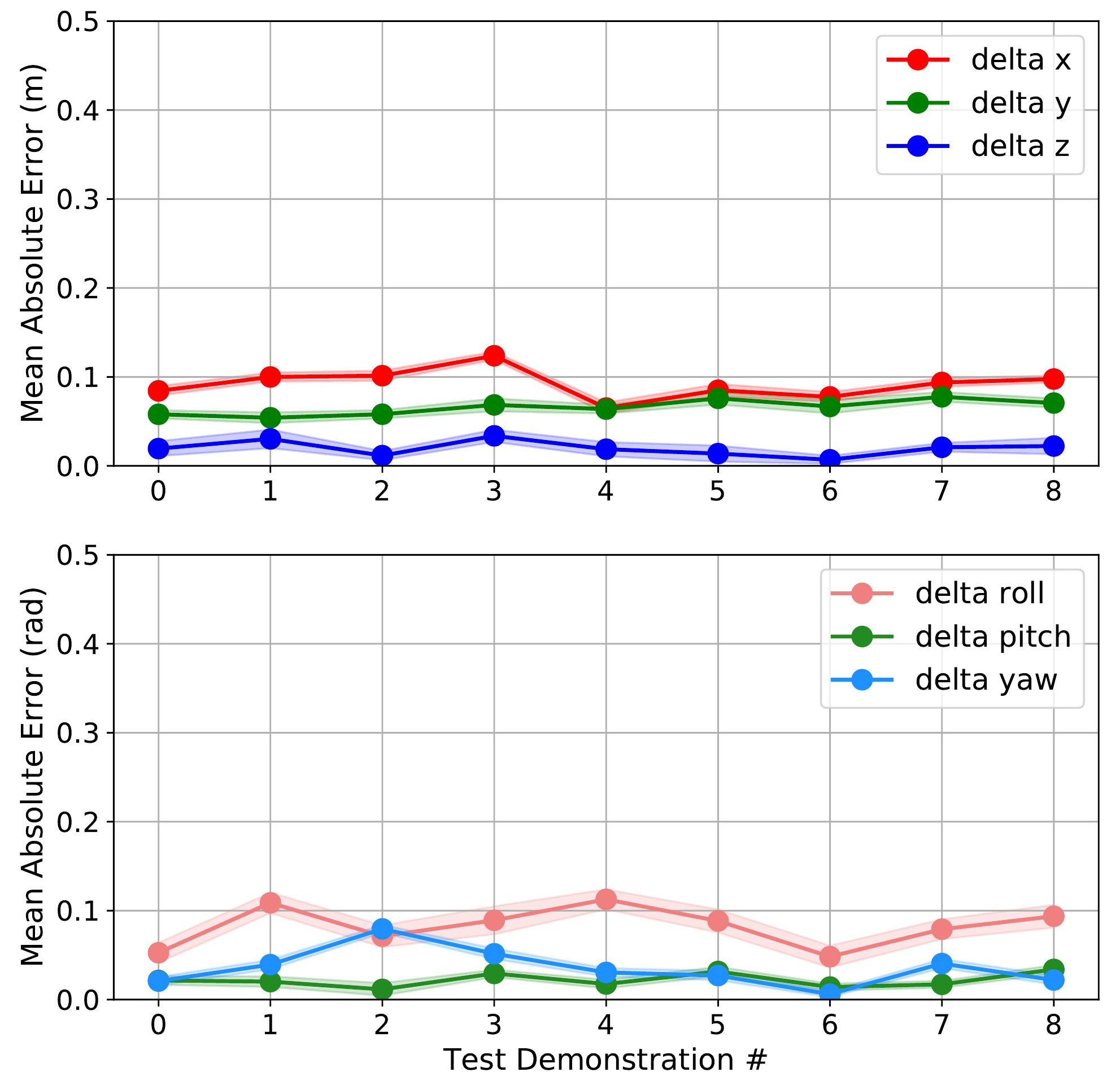}
    \vspace{-1\baselineskip}
    \caption{}
    \end{subfigure}
    \vspace{-0.5\baselineskip}
    \caption{Mean Absolute Error between inferred poses $\hat{\mathbf{p}}$ and commanded poses $\mathbf{p}$ during teleoperation for \textbf{(a)} placing on, \textbf{(b)} facing cups, \textbf{(c)} placing in. The reported error values are across 10 demonstrations (X-axis) not seen during training.}
    \label{fig:res_all_robot_eef_mae}
\end{figure*}

Lastly we demonstrate that using the learned latent grounding of the taught linguistic symbols we can regress end effector positions which capture the meaning behind the symbol (and its associated task). Figure \ref{fig:res_all_robot_eef_mae} reports the mean absolute error between inferred poses $\hat{\mathbf{p}}$ and the true demonstrated ones $\mathbf{p}$ for all three teleoperated tasks. The plots reflect that for certain tasks the model learns to predict more reliably only along the end effector axes that matter for the success of the task (in the way it has been demonstrated)---e.g. for placing \textit{on} and \textit{in} we get lower error across X/Y/Z as compared to when making the cups \textit{face} each other. Respectively, the \textit{facing} task puts more weight on the Roll and Pitch axes (which matter for the cups to have the right orientation) and less weight on the Yaw or on the translational X/Y/Z axes of the end effector.

\section{Conclusion}
\vspace{-3mm}
\label{sec:conclusion}
Effective human-robot collaboration requires shared task representations that are both interpretable and suitable for task completion. 
We present a framework which allows human demonstrators to teach how to ground high-level spatial concepts in their sensory input. We show that while interpretable to the human, due to the disentanglement we explicitly optimize for, the learned latent space is also useful to tasks downstream. In particular, using photorealistic synthetic data we show how such a feature space can be used by an agent to derive explanations for a set of demonstrations, using the symbols it has been taught a priori. We also show how such discrete symbolic representations can be used as a building block for primitive action policies in the context of a robotic agent performing a table-top manipulation task. Future work will involve applying the explain and repeat framework on tabletop manipulation tasks of compositional nature, together with learning modifiers over the taught spatial symbols - e.g. "\textit{more/less} to the left", etc.



\clearpage
\acknowledgments{This work is partly supported by funding from the Turing Institute, as part of the Safe AI for surgical assistance project}


\bibliography{main}  

\clearpage

\begin{appendices}
\section{Data Processing and Network Architecture}
\label{appendix:architecture}

Preprocessing the gathered data consists of extracting the semantic masks, corresponding to each object in the scene, from the raw RGBD pixel-level channels of information and all object and relational labels associated with each pair of objects in a given scene. As issues of semantic segmentation are not the focus of our work, we start with a system that provides us the semantic masks for each object present in the scene from raw observation. In our  robot experiments, the RGB part of the input is fed to a pre-trained Mask R-CNN model, which dictates the partial labelling afterwards. For the BlocksWorld we can extract the masks deterministically, since we have access to the full state of the scene.

Elementary Dependency Structures (EDS) \cite{eds} and the wide-coverage English Resource Grammar \cite{erg} are used to perform this step \cite{eds,erg}. The resultant \verb![target relations referent]! tuples are used to perform weak labelling over sequence of observations that comprise the demonstration. Errors in the labels produces by the parsing procedure are not expected - the process is deterministic and the parsed NL instructions always follow a predefined template.

For example, if we have \verb![yellow_cube, {left, front} , blue_cube]! as a parsed description and the semantic segmentation model detects a \verb!yellow_cube! and a \verb!blue_cube! present in the input image, this results in a single labelled data point ($\mathbf{x}_i, \mathbf{y}_i, \mathbf{o}_{ti}, \mathbf{o}_{ri}$) being added to $\mathcal{W}$, where $\mathbf{x}_i = \{I_{tar}, I_{ref})$ and $\mathbf{y}_i = \{left, front\}$, $\mathbf{o}_{ti} = \{yellow, cube\}$, $\mathbf{o}_{ri} = \{blue, cube\}$. Any segmented pair whose labels do not appear in the description is added to $\mathcal{W}$ as an unlabelled data point.

The model architecture is implemented in the Chainer  framework\footnote{https://docs.chainer.org/en/stable/}. The encoder network takes as input a set of RGBD  128x128 pixel images, a 128x128 binary segmentation mask, and a set of object and relational labels. It tries to reconstruct the same set of RGBD 128x128 pixel images, masked with the corresponding binary segmentation mask, and predict the all labels which are not \textit{unknown}.

\begin{table}[h]
\begin{subtable}{0.32\linewidth}
{\begin{tabular}{c}
\hline
\hline
\textbf{Encoder}                \\ \hline \hline
FC (2x8) Output LogNormal       \\ \hline
FC (256)                        \\ \hline
Conv (k=3, s=2, p=1, c=64)      \\ \hline
Conv (k=3, s=2, p=1, c=64)      \\ \hline
Conv (k=3, s=2, p=1, c=64)      \\ \hline
Conv (k=3, s=2, p=1, c=32)      \\ \hline
Conv (k=3, s=2, p=1, c=32)      \\ \hline
Input Image {[}128 x 128 x C{]} \\ \hline
\end{tabular}}
\caption{Encoder}\label{tab:1}
\end{subtable}
\begin{subtable}{0.32\linewidth}
{\begin{tabular}{c}
\hline
\hline
\textbf{Decoder}           \\ \hline \hline
Output Logits              \\ \hline
Conv (k=3, s=2, p=1, c=C)  \\ \hline
Conv (k=3, s=2, p=1, c=64) \\ \hline
Conv (k=3, s=2, p=1, c=64) \\ \hline
Conv (k=3, s=2, p=1, c=64) \\ \hline
append coord channels      \\ \hline
tile (128, 128, 8)         \\ \hline
Input Vector {[}8{]}       \\ \hline
\end{tabular}}
\caption{Decoder}\label{tab:2}
\end{subtable}
\begin{subtable}{0.32\linewidth}
{
\begin{tabular}{c}
\hline
\hline
\textbf{Operator}           \\ \hline \hline
FC (2 x 6) Output Lognormal \\ \hline
FC (64)                     \\ \hline
FC (256)                    \\ \hline
Input Vector {[}2 x 8{]}    \\ \hline
\end{tabular}}
\caption{Operator}\label{tab:3}
\end{subtable}
	
    \caption{Network architectures used for the reported models. (a) and (c) are standard convolutional and fully-connected MLP networks, (b) is a spatial broadcast decoder, described in \cite{watters2019spatial}}
\label{table:model_architectures}
\vspace{-3mm}
\end{table}

Across all experiments, training is performed for a fixed number of 50 epochs using a batch size of 32. The dimensionality of the latent space $|\mathbf{Z}|$ = 8 across all experiments. The dimensionality of $|\mathbf{C}|$ = 6 for the BlocksWorld experiments and $|\mathbf{C}|$ = 3 for the robot teleoperation experiments. The Adam optimizer \citep{kingma2014adam} is used through the learning process with the following values for its parameters---($learning rate=0.001, \beta1=0.9, \beta2=0.999, eps=1e-08, weight decay rate=0, amsgrad=False$)

For all experiments, the values (unless when set to 0) for the three coefficients from Equation \ref{eq:loss} are:
\begin{itemize}
\item $\alpha = 1, \beta = 10, \gamma = 50000$
\end{itemize}

The values are chosen empirically in a manner such that all the loss terms have similar magnitude and thus none of them overwhelms the gradient updates while training the full model.
\section{Plan Segmentation Elaboration}
\label{appendix:plan_segmentation}

\begin{algorithm}[H]
\caption{Movement Prescription Seq Identification}
\label{alg:plan_extraction}
	\KwIn{Sequence of $T$ observations $\mathcal{I} = \{\mathbf{I}_1 \ldots \mathbf{I}_T\}$}
	\KwIn{Referent object labels of $\mathbf{o}_{ref}$}
    \KwIn{Encoder network $q_{\theta}$, $\Sigma_{stationary}, \Sigma_{moving}$}
    \KwOut{Movement prescription sequence $\hat{\mathcal{S}}$} \BlankLine
    $\hat{\mathcal{S}} = []$\;
    $O_{tar} \leftarrow$ segment($\mathcal{I}, \mathbf{o}_{ref}$)\;
    For every two objects extract all tuples $\{(\mathbf{X},\mathbf{Y}, \mathbf{o}_1, \mathbf{o}_2)\} \leftarrow$ preproc($\mathcal{I} | {O_{tar} \bigcup \mathbf{o}_{ref} }$)\;
    \For{each object pair in $\{( \mathbf{o}_{tar},  \mathbf{o}_{ref}) | \mathbf{o}_{tar} \in O_{tar}\}$}{
        $\mathcal{I}_{mask} \leftarrow \{(\mathbf{x},\mathbf{y},  \mathbf{o}_{tar}, \mathbf{o}_{ref}) \in (\mathbf{X},\mathbf{Y}) | \mathbf{o}_{tar} \in \mathbf{o}_1$ \& $\mathbf{o}_{ref} \in \mathbf{o}_2\}$)\;
        $\mathcal{T} \leftarrow q_{\theta}(\mathcal{I}_{mask})$\; 
        $\hat{\mathbf{s}} \leftarrow []$\;
        \For{each ($\tau_t, \tau_{t+1}$ in zip($\mathcal{T}[:-1], \mathcal{T}[1:]$)}{
        \If{$\mathcal{N}(\tau_{t+1}|\tau_t,\Sigma_{mov}) > \mathcal{N}(\tau_{t+1}| \tau_t,\Sigma_{stat})$}{
            Append $ \mathbf{o}_{tar}$ to $\hat{\mathbf{s}}$\;
        }
        \Else{
            Append $\emptyset$ to $\hat{\mathbf{s}}$\;
        }
        }
    Append $\hat{\mathbf{s}}$ to $\hat{\mathcal{S}}$\;
    }
    return $\hat{\mathcal{S}}$\;
\end{algorithm}

\begin{figure*}[h]
\centering
\includegraphics[width=1\linewidth]{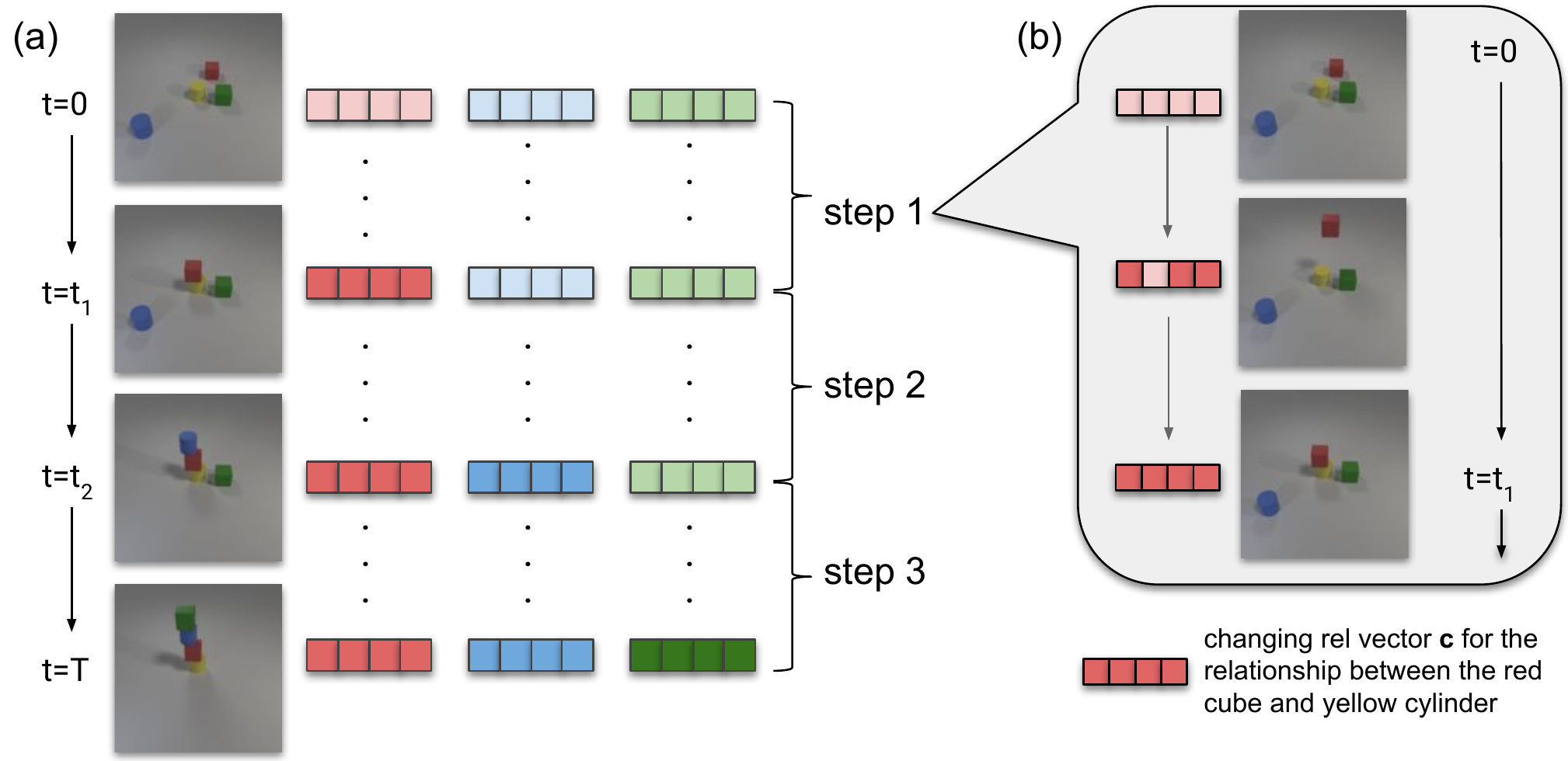}
\caption{Plan segmentation pipeline.\textbf{(a)} Infer a movement prescription sequence---\textit{what} is moved after what---and \textbf{(b)} infer \textit{how} is each object moved when it is manipulated. In this example the red, green and blue object are sequentially stacked on top of the yellow one. A change in the color shade corresponds to \textbf{(a)} an object being moved or \textbf{(b)} an object changing the way it relates to the reference object in the scene along one or more concept groups.}
\label{fig:plan_segmentation}
\end{figure*}

As described in sections \ref{sec:explain_n_repeat} and \ref{sec:infer_symbol_plans}, ee use the trained model $q_{\theta}$ to convert a sequence of raw observations $\mathcal{I}$---images in Figure \ref{fig:plan_segmentation} (a)---into a trace of $T$ relational embeddings $\mathcal{T} = \{\tau_1 \ldots \tau_T\}, \tau_i \in \mathbb{R}^C$---colored blocks in Figure \ref{fig:plan_segmentation}. In order to detect whether the two objects move with respect to each other, a likelihood ratio test with two normal distributions---$\mathcal{N}_{stationary}$ and $\mathcal{N}_{moving}$---is performed on every two sequential embeddings $\tau_t$ and $\tau_{t+1}$. For the purpose of the experiments, both $\Sigma_{mov}$ and $\Sigma_{stat}$ are diagonal covariance matrices with $\sigma_{ii}$ being 1 and 0.1 respectively. More details can be found in Algorithm \ref{alg:plan_extraction} above and Figure \ref{fig:movement_prescription_sequence} below.

Additionally, for each part of a given trace $\mathcal{T}$ where the objects are moving with respect to each other, we can use the parametrised distributions $K$ for each cluster in each group in $\mathbb{R}^C$ (including ones for \textit{unlabelled} relationships) for an additional likelihood-ratio test to decide how the objects move---see Figure \ref{fig:plan_segmentation} (b). The latter is equivalent to essentially checking when and object changes membership along each concept group with respect to the reference object in the scene. This allows us to go from a sequence of observations $\mathcal{I}_{mask}$---masked images in Figure \ref{fig:movement_prescription_sequence}---to what is essentially a symbolic plan $\mathcal{Y}$.

It is noted that such an approach might capture \textit{noisy} steps that do not represent the intent of the demonstrator---e.g. we move an object from being left to right with respect to another object by going behind it in the intermediate states. The upper-described procedure would infer that the moved object being behind the static one is a valid substep when that is not actually part of the user's intent. Thus, in the presence of more demonstrations, we filter steps from the plan that are not identified in all demonstrations, in order to produce the essence of the demonstrated task. The goal is to try to identify the most invariant \textit{plan} that best explains a set of demonstrations that have the same underlying goal---e.g. if we have two demonstrations where an object is moved from left to right with respect to another static object, we aim to identify an explanation that ignores the fact that once we move in front and once behind that object.


\begin{figure*}[h]
\centering
\includegraphics[width=1\linewidth]{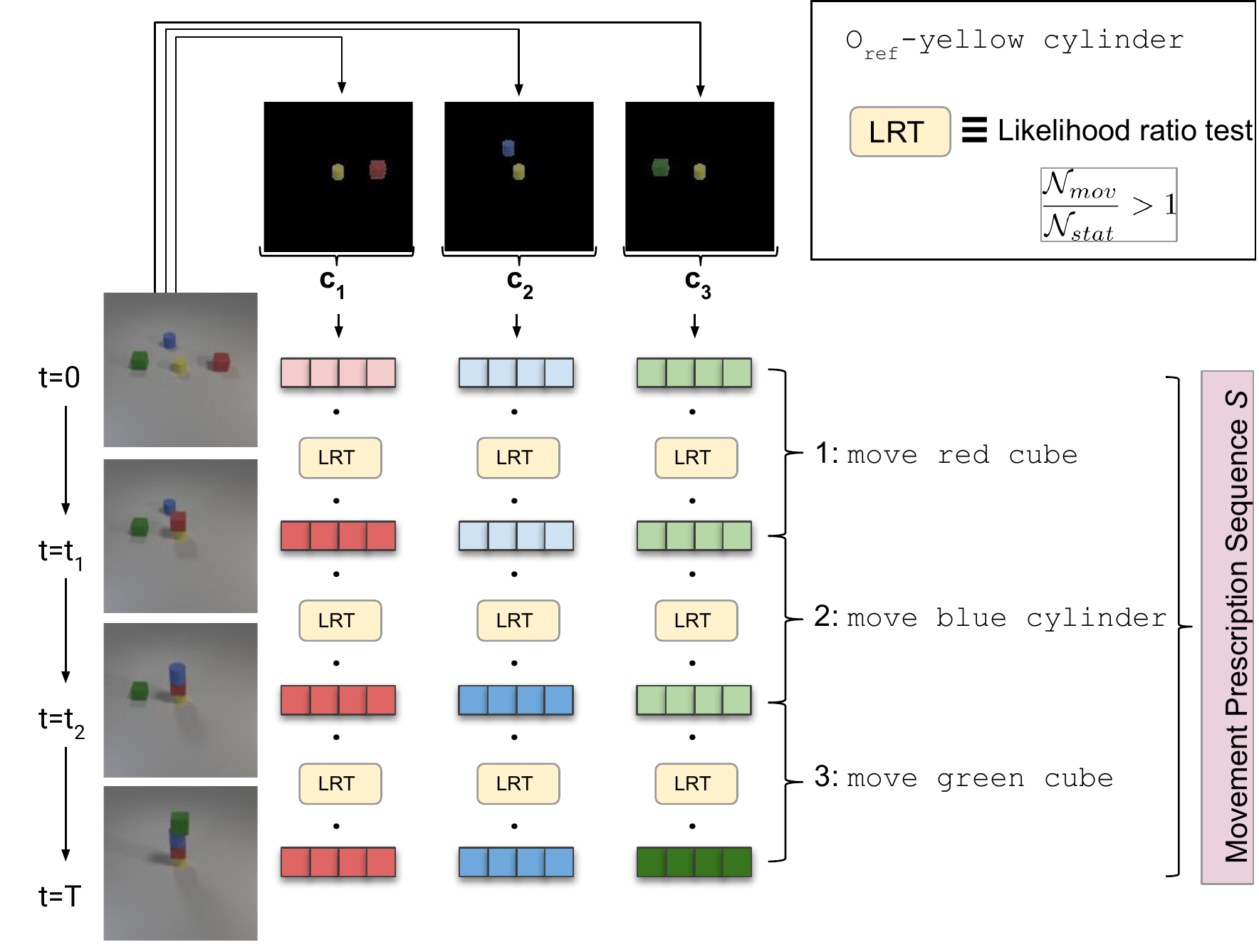}
\caption{Visual illustration of the movement prescription sequence procedure described in algorithm \ref{alg:plan_extraction} and Figure \ref{fig:plan_segmentation} (a)}
\label{fig:movement_prescription_sequence}
\end{figure*}
\section{Disentanglement Analysis of the Information Bottleneck $\mathbf{C}$}
\label{appendix:violins}

\begin{figure*}[h]
    \centering
    \includegraphics[width=0.9\linewidth]{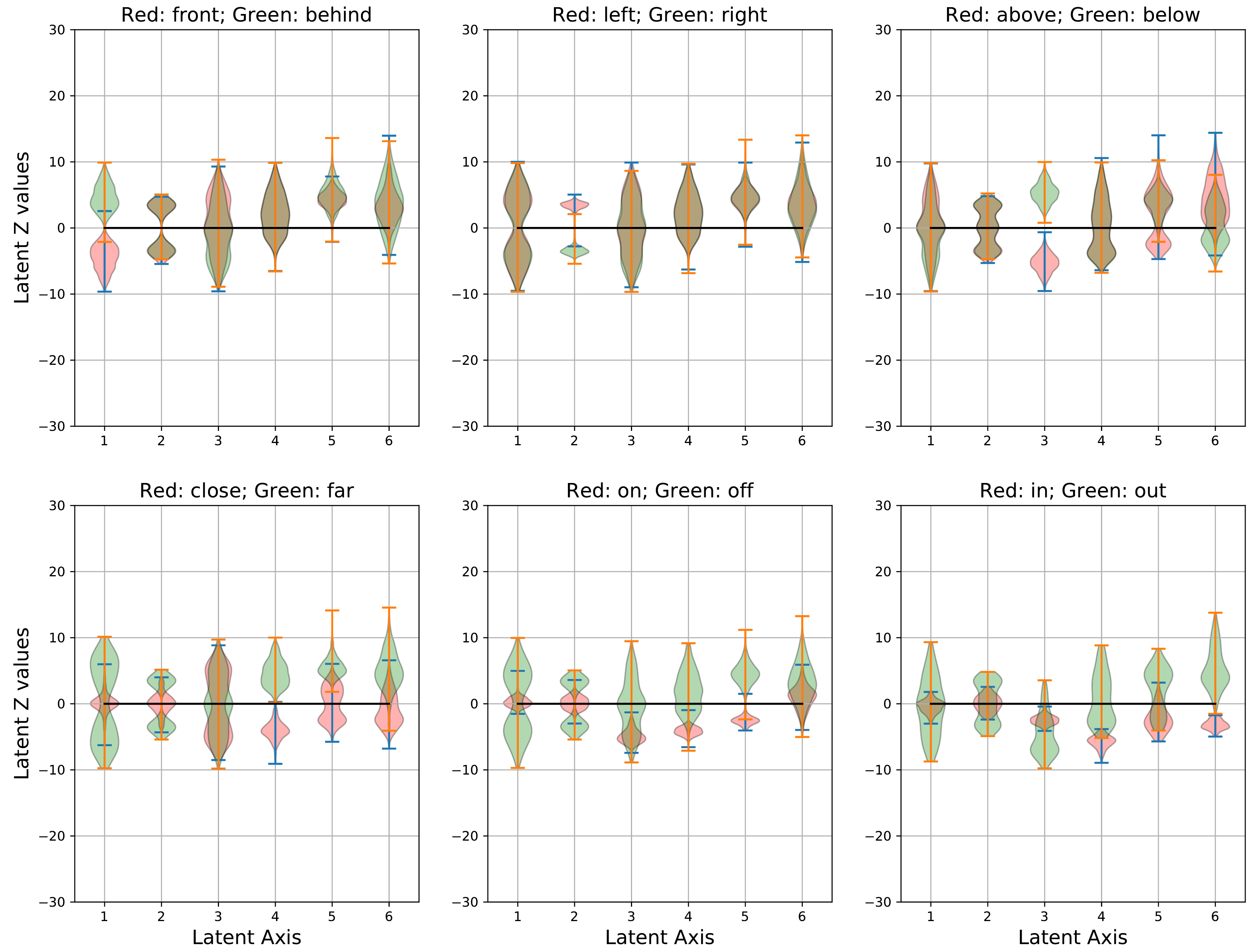}
    \caption{Evaluation of the degree of disentanglement in the latent space $\mathbf{C}$ for each concept group across the different baseline models used in the ablation study --- No $\mathcal{R}$, No $\mathcal{Q}_{obj}$}
    \label{fig:violin_plots_1}
\end{figure*}

In order to bring additional clarity in the properties of the learned latent relational space, we provide violin plots for the distributions of data points from each concept group (X axis on each plot). We can observe that model which do not utilise object label information in training the object embeddings $\mathbf{z}$---Figure \ref{fig:violin_plots_1} and Figure \ref{fig:violin_plots_3}---tend to learn relational embeddings $\mathbf{c}$ which fall in a tighter region, centered around 0, due to the influence of the KL objective. We hypothesise that this is one of the reasons for these models to underperform in inferring the movement prescription sequence for the given demonstrations. With the latent clusters being projected closer, tuning the parameters of the distributions used in the movement likelihood ratio test might required.

\begin{figure*}[h]
    \centering
    \includegraphics[width=0.9\linewidth]{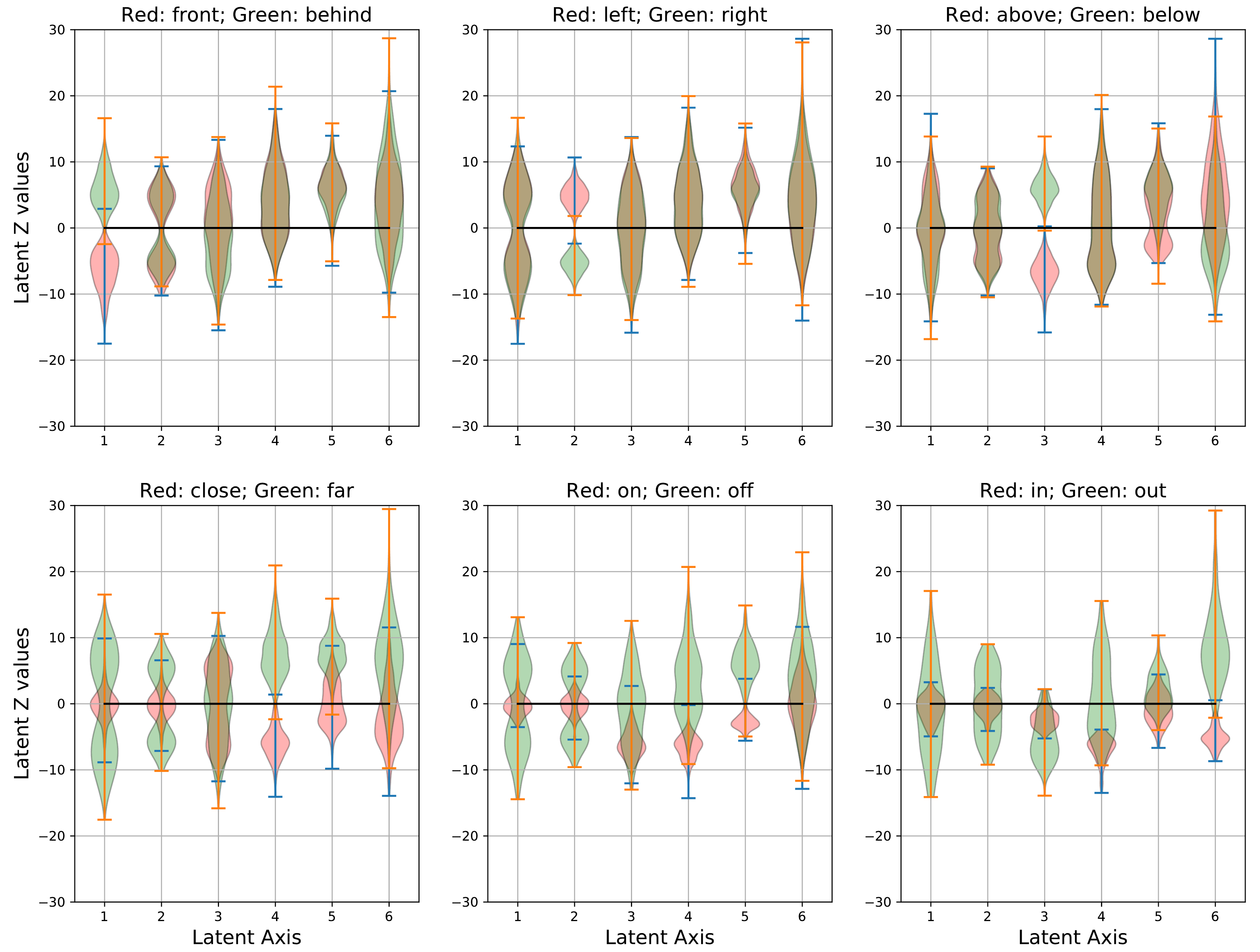}
    \caption{Evaluation of the degree of disentanglement in the latent space $\mathbf{C}$ for each concept group across the different baseline models used in the ablation study --- No $\mathcal{R}$, With $\mathcal{Q}_{obj}$}
    \label{fig:violin_plots_2}
\end{figure*}
\begin{figure*}[h]
    \centering
    \includegraphics[width=0.9\linewidth]{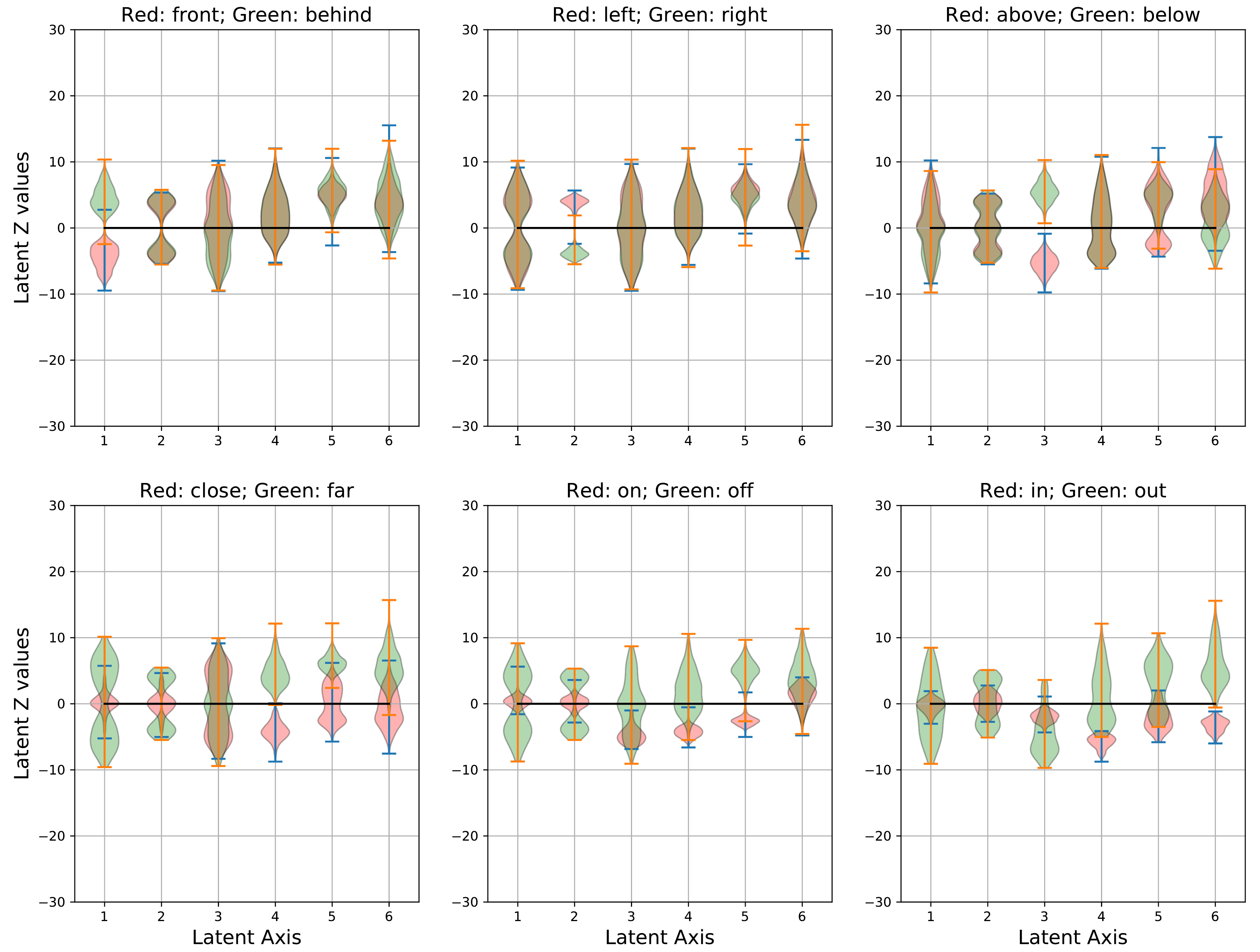}
    \caption{Evaluation of the degree of disentanglement in the latent space $\mathbf{C}$ for each concept group across the different baseline models used in the ablation study --- With $\mathcal{R}$, No $\mathcal{Q}_{obj}$}
    \label{fig:violin_plots_3}
\end{figure*}
\begin{figure*}[h]
    \centering
    \includegraphics[width=0.9\linewidth]{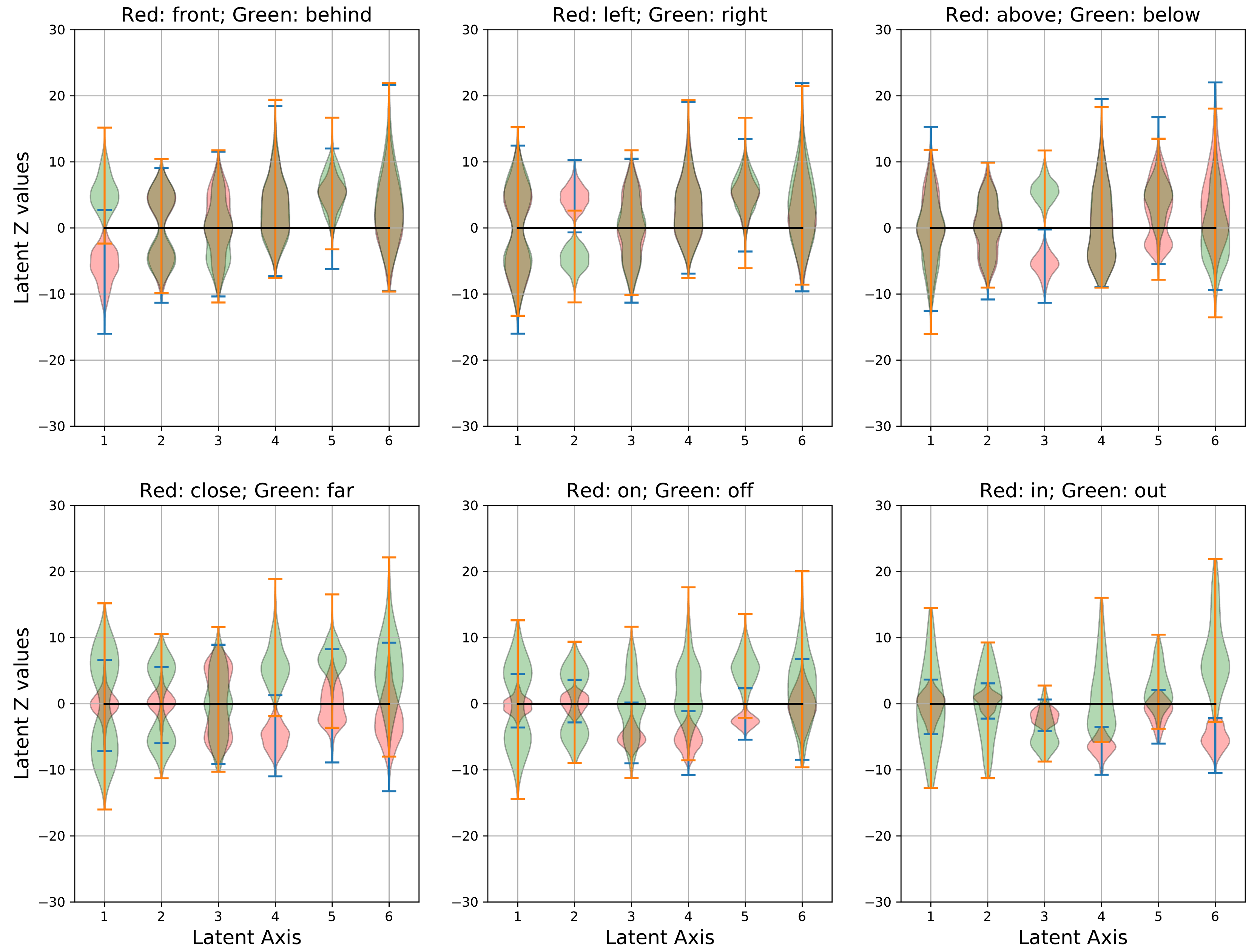}
    \caption{Evaluation of the degree of disentanglement in the latent space $\mathbf{C}$ for each concept group across the different baseline models used in the ablation study --- With $\mathcal{R}$, With $\mathcal{Q}_{obj}$}
    \label{fig:violin_plots_4}
\end{figure*}
\end{appendices}

\end{document}